\documentclass[letterpaper, 10 pt, conference]{ieeeconf}  %
\usepackage{times}
\usepackage{epsfig}
\usepackage{graphicx}
\usepackage{amsmath}
\usepackage{amssymb}
\usepackage{soul}
\usepackage{multirow, booktabs}
\usepackage{siunitx}
\usepackage{makecell}
\usepackage{xcolor}
\usepackage{array,multirow,graphicx}
\usepackage{mathtools}

\usepackage{xcolor}
\usepackage{bbm}
\usepackage{arydshln}
\usepackage{tabularx}

\usepackage[linesnumbered,ruled,vlined]{algorithm2e}
\usepackage{hyperref}

\hypersetup{colorlinks,linkcolor={blue},citecolor={green}} 
\IEEEoverridecommandlockouts                              %

\newcommand{\sota}{state-of-the-art }
\newcommand{\wildtrack}{WildTrack }
\newcommand{\multiviewx}{MultiViewX }

\newcommand{\etal}{\textit{et al.}}

\title{\LARGE \bf
Bringing Generalization to Deep Multi-View Pedestrian Detection
}

\author{Jeet Vora$^{1}$, Swetanjal Dutta$^{1}$, Kanishk Jain$^{1}$, Shyamgopal Karthik$^{2}$, Vineet Gandhi$^{1}$ %
\thanks{$^{1}$ CVIT, KCIS, International Institute for Information Technology, Hyderabad, India
        {\tt\footnotesize jeet.vora@research.iiit.ac.in}}%
 \thanks{$^{2}$University of T{\.u}bingen, Germany}%
}

\begin{document}

\maketitle
\thispagestyle{empty}
\pagestyle{empty}

\begin{abstract}

Multi-view Detection (MVD) is highly effective for occlusion reasoning in a crowded environment. While recent works using deep learning have made significant advances in the field, they have overlooked the generalization aspect, which makes them \emph{impractical for real-world deployment}. The key novelty of our work is to \emph{formalize} three critical forms of generalization and \emph{propose experiments to evaluate them}:  generalization with i) a varying number of cameras, ii) varying camera positions, and finally, iii) to new scenes. We find that existing state-of-the-art models show poor generalization by overfitting to a single scene and camera configuration. To address the concerns: (a) we propose a novel Generalized MVD (GMVD) dataset, assimilating diverse scenes with changing daytime, camera configurations, varying number of cameras, and (b) we discuss the properties essential to bring generalization to MVD and propose a barebones model to incorporate them. We perform a comprehensive set of experiments on the WildTrack, MultiViewX and the GMVD datasets to motivate the necessity to evaluate generalization abilities of MVD methods and to demonstrate the efficacy of the proposed approach. The code and the proposed dataset can be found at \url{https://github.com/jeetv/GMVD}

\end{abstract}

\section{Introduction}

\begin{minipage}{0.45\textwidth}
``Essentially all models are wrong, but some are useful.''
\end{minipage}
\\[3pt]
\rightline{{\rm --- George E. P. Box}}

In this work, we pursue the problem of Multi-View Detection (MVD), a mainstream solution for dealing with occlusions, especially when detecting humans/pedestrians in crowded settings. The input to MVD methods is images from multiple calibrated cameras observing the same area from different viewpoints with an overlapping field of view. The predicted output is an occupancy map~\cite{Fleuret2008MulticameraPT} in the ground plane (bird's eye view). The solutions of MVD has evolved from classical methods~\cite{Fleuret2008MulticameraPT,berclaz2011multiple,alahi2011sparsity}, to hybrid approaches~\cite{kong2020foveabox} to end-to-end trainable deep learning architectures~\cite{hou2020multiview}. Expectedly, the current landscape of MVD is dominated by end-to-end trainable deep learning methods~\cite{hou2020multiview,hou2021multiview,song2021stacked}. We argue that by \emph{training and testing on homogeneous data}, current deep MVD methods have overlooked critical fundamental concerns, and to render them \emph{useful}, the focus should shift towards their generalization abilities.

\begin{figure}[t]
    \centering
    \includegraphics[width=\linewidth]{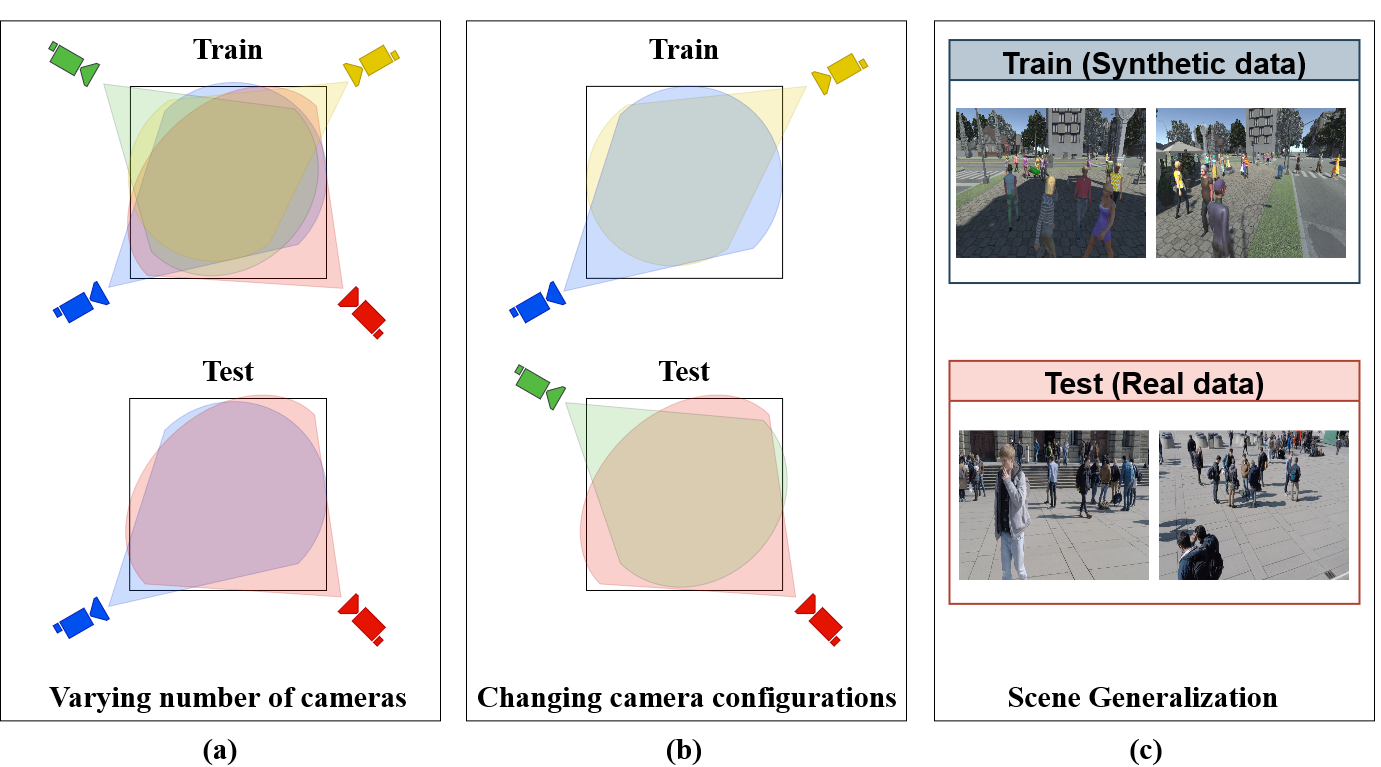}
    \caption{Three forms of generalization required in MVD: (a) varying number of cameras, (b) different camera configurations, and (c) generalizing to new scenes.}
    \label{fig:generalization}
\end{figure}

 Ideally, three forms of generalization abilities are essential for the practical scalability and deployment of MVD methods, which is illustrated in Fig.~\ref{fig:generalization}:
\begin{enumerate}
    \item \emph{Varying number of cameras:} The model should adapt to a varying number of cameras (a network trained on six camera views, should work on a setup with five cameras). 
    \item \emph{Varying configuration:} The model should not overfit to the specific camera configuration. The performance should be similar even with altered camera positions, as long as they span the dedicated area. 
    \item \emph{Varying scenes:} Models trained on one scene should work on another (model trained on a traffic signal should work on a setup inside a university). 
\end{enumerate}
Surprisingly, the existing deep learning-based MVD methods are primarily trained and tested with the same camera configuration, on the same scene, using the same number of cameras. Even the environmental conditions (time, weather, etc.) are similar across train and test splits. For instance, the most commonly used Wildtrack dataset~\cite{Chavdarova2018WILDTRACKAM} includes a 200 second recording from all cameras, where the first 3 minutes are used for training and the rest of the 20 seconds are used for testing. We argue that the current State Of The Art (SOTA) methods are seriously hindered from the deployment perspective. The current models~\cite{hou2020multiview,hou2021multiview,song2021stacked} will break if a camera malfunctions. They will need retraining if a camera needs to be added to the setup. Furthermore, our experiments show that the performance significantly drops if the camera positions or the scene is varied. The SOTA models also seem to overfit to the order in which the cameras are sent to the model.

 \begin{figure}
    \centering
    \includegraphics[width=\linewidth]{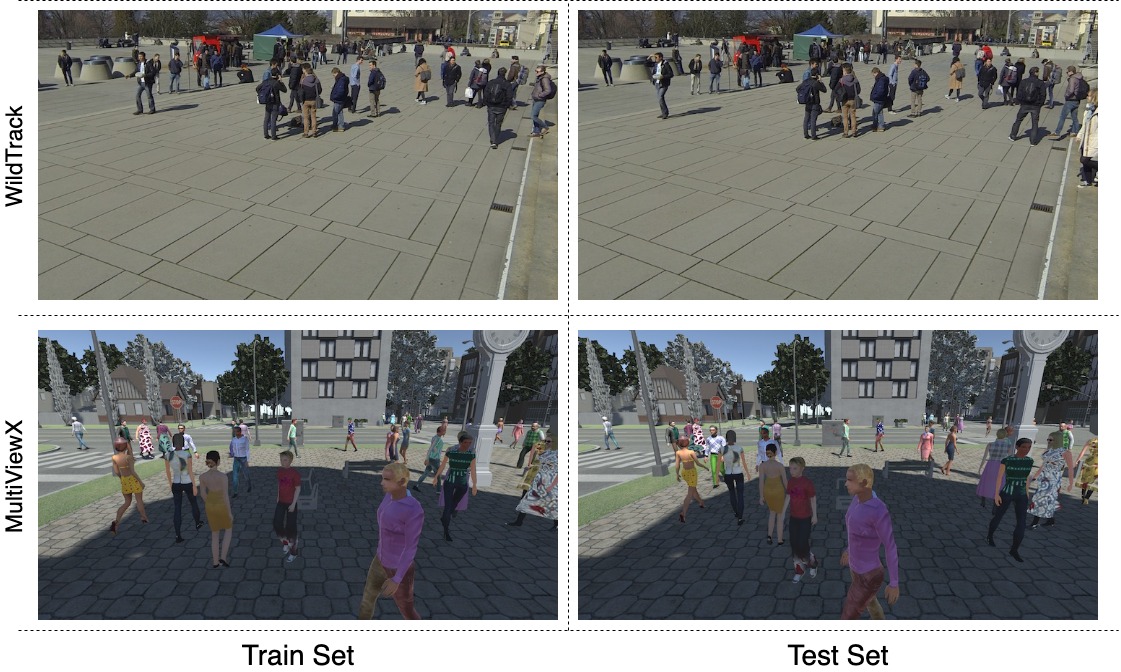}
    \caption{The train and test sets of Wildtrack (first row) and MultiViewX datasets (second row) have significant overlap. We show the last image of the training set (left) and the first image of the test set (right). In both datasets, the appearance of several pedestrians is already seen in the training set. In Wildtrack, there are many static pedestrians as well.}
    \label{fig:dataset_issues}
\end{figure}

The absence of a diverse dataset is a major shortcoming. The available datasets: Wildtrack (real) and MultiViewX (synthetic), comprise a single short sequence, where initial frames are used for training and later for testing. In Figure~\ref{fig:dataset_issues}, we show that the evaluation strategy in both datasets is unreliable and prone to overfitting. To this end, we propose a novel Generalized MVD (GMVD) dataset. Given the privacy concerns, COVID restrictions, hardware setup difficulties, the requirement of manual annotations, etc., we believe curating a sizeable synthetic dataset is the right way forward. Henceforth, we use Unity and the GTA game environment to capture the GMVD dataset. It includes about 53 sequences captured in 7 different scenes with significant variations in camera configuration, weather, lighting conditions, pedestrian appearance, etc. The number of cameras also varies across scenes. We use 6 scenes for training and 1 scene for testing. The proposed GMVD dataset sets up a new benchmark for evaluating MVD with generalization. It further allows reserving valuable real-world footages~\cite{Chavdarova2018WILDTRACKAM} directly for testing. 

Furthermore, we suggest a set of design guidelines to ensure practical usability of Deep MVD methods. We demonstrate that permutation invariance, transfer learning, and regularization are vital for generalization. We improve the baseline architecture~\cite{hou2020multiview} with appropriate changes and establish SOTA generalization for MVD. We want to emphasize that we do not claim any major architectural novelty, and our work focuses on the barebone baseline architecture. Overall, our work makes the following contributions:

\begin{enumerate}
    \item We conceptualize and emphasize the importance of generalization in MVD and propose a novel GMVD dataset for the same.
    \item We highlight the shortcomings of the current evaluation methodology and propose novel experimental setup on existing datasets. 
    \item We adapt the baseline architecture to bring generalization to deep MVD. We show that permutation invariance is crucial for MVD and average pooling is one minimal way to achieve it. We propose a novel drop view regularization.
    \item  We back our claims using an extensive set of experiments and ablation studies. We show staggering improvements in scene and configuration generalization, paving the way for a practicable MVD.
\end{enumerate}

\section{Related Work}

\subsection{Classical Methods}
Seminal work by Fleuret \etal~\cite{Fleuret2008MulticameraPT} cast MVD as predicting occupancy probabilities over a discrete grid, an idea which has stood the test of time. The classical methods in MVD rely on background subtraction to compute likelihood over a fixed set of anchor boxes derived using scene geometry, project them on the top view and adopt conditional random field (CRF) or mean-field inference for spatial aggregration~\cite{Fleuret2008MulticameraPT,berclaz2011multiple,alahi2011sparsity}. The classical methods, however observe a gradual degradation in detection performance with increased crowds, as the background subtraction becomes less effective with increase in crowds and clutter. Some methods do away with background subtraction and rely on handcrafted classifiers~\cite{Roig2011ConditionalRF} instead.

 \begin{figure*}[!t]
    \centering
    \includegraphics[width=\linewidth]{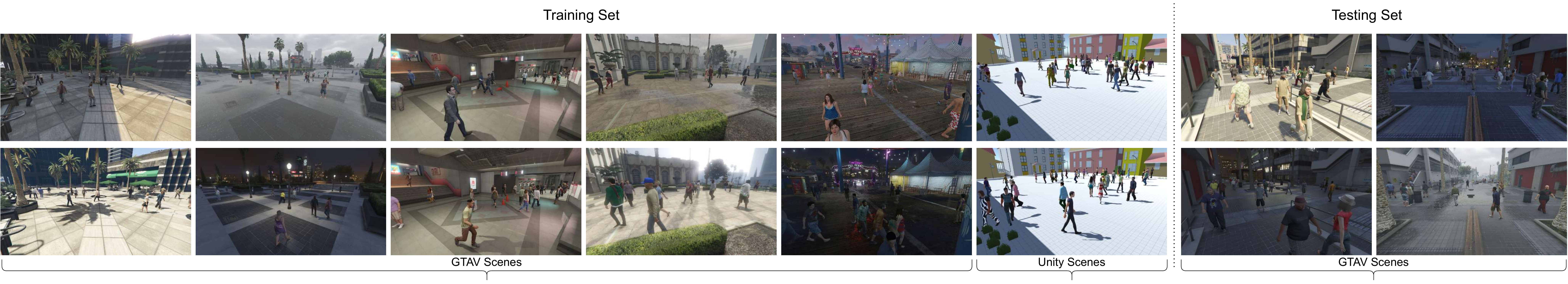}
    \caption{The proposed GMVD Dataset includes seven scenes. Each column illustrates frames from one of the views from two different sequences of the same scene. The first six scenes are used for training and the last scene with two configurations are reserved for testing. Additionally, there are noticeable lighting and weather variations within each scene.}
    \label{fig:dataset}
\end{figure*}

\begin{table*}
\centering
\small
\caption{Dataset Statistics for various MVD datasets. Our proposed GMVD dataset is the largest and most diverse dataset on a variety of metrics. Avg. coverage refers to the average number of cameras that cover each point on the ground plane. }
\resizebox{\textwidth}{!}{%
\begin{tabular}{@{}lcccccccccc@{}}
\toprule
Dataset  & Track Labels & IDs  & \# Scenes & \# Training Frames & \# Testing Frames & \# Cameras & \# Sequences & Avg. Coverage  \\ 
\midrule
WildTrack & \checkmark & 313 & 1         & 360  & 40 & 7 & 1 & 3.74            \\
MultiViewX & \checkmark & 350 & 1 & 360  & 40 & 6 & 1 & 4.41            \\
GMVD (Ours) & \checkmark & 2800 & 7 & 4983 & 1012  & {3, 5, 6, 7, 8} & 53  & {2.76 - 6.4}   \\ \bottomrule
\end{tabular}%
}

\label{tab:dataset-stats}
\end{table*}

\subsection{Anchor based MVD}
Anchor based MVD methods replace background subtraction with anchor-based deep pedestrian detectors like Faster R-CNN~\cite{Ren2015FasterRT}, SSD~\cite{Liu2016SSDSS} and YOLO~\cite{Redmon2016YouOL}. Some of these methods process each view separately~\cite{xu2016multi} and some process them simultaneously~\cite{Baqu2017DeepOR, Chavdarova2017DeepMP}. The inaccuracies in the pre-defined anchor boxes~\cite{kong2020foveabox} limit the performance of these methods. Even if the boxes are correct, locating the exact ground point to project in each 2D bounding box presents a challenge and leads to a significant amount of errors. Moreover, some of the Anchor based methods still rely on operations outside of Convolutional Neural Networks (CNNs), requiring to work out a balance between different potential terms~\cite{Baqu2017DeepOR}.

\subsection{End-to-end Deep MVD}

MVDet \cite{hou2020multiview} is a recent anchor-free approach that aggregates multi-view information by perspective transformation and concatenating multi-view feature map onto the ground plane and then performs large kernel convolution for spatial aggregation. It overcomes limitations of manual tuning of CRF potentials, reliance on pre-defined 3D anchor boxes and projection errors from monocular detectors. It aggregates projected features from a ResNet~\cite{He2016DeepRL} backbone using three convolutional layers to predict the final occupancy map. MVDet achieves notable improvement over the preceding anchor based methods (over 14\% improvement on the \wildtrack dataset~\cite{Chavdarova2018WILDTRACKAM}). The idea from \cite{hou2020multiview} was further enhanced by using deformable transformers~\cite{zhu2020deformable} to improve the feature aggregation in MVDeTr~\cite{hou2021multiview}. More recently, SHOT~\cite{song2021stacked} introduced a combination of homographies at multiple heights to improve the quality of the projections.

\section{Proposed Dataset}

We propose a new MVD dataset incorporating the three forms of generalization discussed above (Figure~\ref{fig:generalization}). Some example frames from the proposed Generalized Multi-View Detection (GMVD) dataset are illustrated in Figure~\ref{fig:dataset}. The GMVD dataset contains diverse non-overlapping scenes within and across training and test sets. In contrast, the existing MVD datasets Wildtrack and MultiViewX include noticeable overlap across train and test splits (single scene, pedestrians appearance, and location), encouraging existing MVD methods to overfit the dataset-specific aspects and thus hindering their practicality. The GMVD dataset, by its design, prevents overfitting from happening by keeping a clear separation in train and test splits.

Capturing a real-world MVD dataset is difficult, primarily because of privacy concerns. The COVID restrictions also restrict crowded human capture. Moreover, such a dataset requires significant manual annotation effort. Consequently, we curate the GMVD dataset using synthetic environments. The GMVD dataset is curated using Grand theft Auto V (GTAV) and Unity Game Engine. We employ two different environments to avoid overfitting to a single synthetic data generation source. This reasoning is aligned with recent works \cite{gong2021mdalu, Zhao_2021_CVPR} which utilize multi-source datasets to improve generalization performance.
The GMVD dataset includes seven distinct scenes, one indoor (subway) and six outdoors. One of the scenes are reserved for the test split. We vary the number of total cameras in each scene and provide different camera configurations within a scene.

 \begin{figure*}[t]
    \centering
    \includegraphics[width=\linewidth]{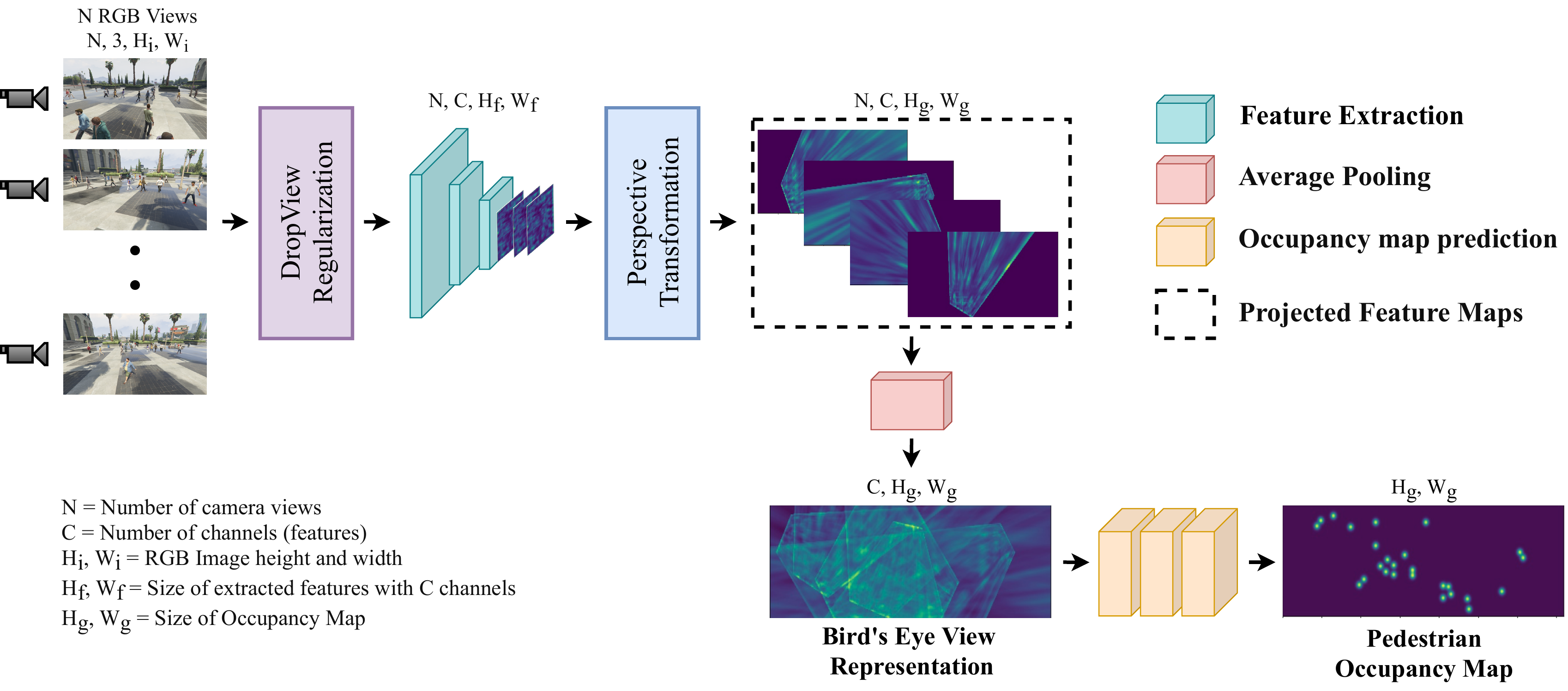}
    \caption{Our proposed architecture: ResNet features are extracted from the input views, which are then projected to the top view. Following this, the projected features across views are pooled and then the final occupancy map is predicted. The use of average pooling across views is crucial in ensuring that our proposed architecture can work for an arbitrary number of views. }
    \label{fig:architecture}
\end{figure*}

Additional salient features of GMVD include daytime variations (morning, afternoon, evening, night) and weather variations (sunny, cloudy, rainy, snowy). We generate multiple short sequences for each scene while randomly varying the daytime and the weather. The generation of multiple random sequences ensures diversity, as different pedestrians (with different clothing and appearance) are picked in each case. The dataset also includes significant variations in lighting conditions. Local illumination sources come into play due to the presence of indoor and night scenes. We compare our dataset with the existing ones in Table \ref{tab:dataset-stats}. Avg. Coverage represents the average amount of cameras observing each location. For GMVD, avg. coverage varies from 2.76-6.4 cameras depending on the scene. In addition to the discussed variations, GMVD is advantageous due to the dataset size, especially in terms of the total number of individual sequences. 

Thereby, we propose the GMVD dataset as a new benchmark for MVD. We further encourage future methods to train on the GMVD dataset and test their performance on sparsely available, difficult to capture real-world datasets like \wildtrack.

\textbf{Dataset Generation:} We used Script Hook V \cite{scripthook} library to interface with the GTAV environment. For each scene, camera positioning and orientation were determined manually so as to increase the camera coverage. All the cameras were positioned above the humans’ average height. Due to hardware limitation, it is commonplace to have a small synchronization delay in real-world multi-camera setups. To emulate such realistic scenario, we induce a small synchronization error (20-100 ms) between different camera views~\cite{Kohl_2020_CVPR_Workshops}. A ground plane was defined for each location, partially overlapping with each camera’s field of view. Only pedestrians inside the ground plane were considered for multi-view detection. We relied on the GTA’s navigational AI engine to avoid collision and to obtain realistic pedestrian behavior. 

In Unity environment, the scene is manually curated by putting together 3d models of street, buildings and other props. We used the PersonX \cite{sun2019dissecting} 3d human models to create the pedestrians. To avoid collision errors (which are present in MultiViewX dataset), pedestrians were spawned at random locations within the region of interest, for every frame.

Since both the environments are synthetic, the 3D-2D correspondences were directly available from the game engines. We use similar procedure as \cite{hou2020multiview} for camera calibration.

\textbf{Track Labels:} Our work focuses on a comprehensive analysis of the problem of Multi-View Detection. However, the proposed dataset can also be useful for the task of multi-view pedestrian tracking. To this end, for the sequences generated from the GTAV environment, we collect the track labels while capturing the data. While we do not use track labels in this work, we provide them with the dataset, which will be beneficial for the community in the future. We provide a total of 125000 frames with track labels. The GTAV frames for the GMVD dataset are regularly sampled from these densely annotated sequences.

\section{Proposed Method}
\label{sec:method}
 We propose an anchor free deep MVD method along the lines of ~\cite{hou2020multiview,hou2021multiview,song2021stacked} specifically tailored to improve the generalization abilities by modifying the training objective and making use of an average pooling strategy on the projected feature maps. The overall architecture is shown in Fig.~\ref{fig:architecture}. The input to our pipeline are multiple calibrated RGB cameras with overlapping fields of view, and the expected output is the occupancy map for pedestrians.
\subsection{Feature Extraction and Perspective Transformation}
\textbf{Feature Extractor:} We use a ResNet18~\cite{He2016DeepRL} backbone as a feature extractor replacing last three strided convolutions with dilated convolutions to have a high spatial resolution of the feature maps. Given $N$ camera views of image size $(3, H_i, W_i)$, where $H_i$ and $W_i$ corresponds to height and width of images, $C$-channel features are extracted for $N$ camera views which corresponds to size $(N, C, H_f, W_f)$, where $H_f$ and $W_f$ represents the  height and width of the extracted features.

\textbf{Perspective Transformation:} The extracted features from the feature extractor are then projected onto the ground plane using a perspective transformation, where $(H_g, W_g)$ corresponds to the height and width of the ground plane grid. Considering the calibrated cameras, \textbf{$K$} represents the intrinsic camera parameters and \textbf{$[R | t]$} represents the extrinsic camera parameters (\textbf{$R$} is the rotation matrix and \textbf{$t$} is the translation vector). 

In the world coordinate system, the ground plane corresponds to \textbf{$ Z = 0 $}, i.e., \textbf{$W = (X, Y, 0, 1)^T$}. A pixel of an image \textbf{$ I= (x,y)^T $} is transformed to the ground plane as follows:
\begin{align}
I = s \begin{pmatrix} x\\ y\\ 1\end{pmatrix} &= K[R|t]\begin{pmatrix} X\\ Y\\ Z\\ 1\end{pmatrix} = P \begin{pmatrix} X\\ Y\\ Z\\1\end{pmatrix}
\end{align}
where \textbf{$s$} is a scaling factor and \textbf{$P$} is a perspective transformation matrix.

\subsection{Spatial Aggregation}
\textbf{Average Pooling:} We first project the ResNet feature maps from each viewpoint on to the bird's eye view using the perspective transformation to obtain the projected feature maps $fm_{i}$ $(\text{where},  i = 1,2,..., N)$. 
Following this, we average pool the projected feature maps \textbf{$fm_{i}$}  to obtain the final bird's eye view feature representation $F$ of size $(C, H_g, W_g)$, which is written as,
\begin{align}
F = \frac{\sum_{i=1}^{N}fm_{i}}{N}.
\end{align}

While there can be many other alternatives to average pooling, we opt for this solution, primarily because it is permutation-invariant. Unlike MVDet, where the camera views ideally need to be input in the same order as training during inference, our proposed solution can accept arbitrary number of views in an arbitrary order. Furthermore, the average pooling solution is free from any learnable parameters which ensures that there is no overfitting introduced due to this operation.
The projected feature maps for $N$ cameras of size $(N, C, H_g, W_g)$ after average pooling, reduces to $(C, H_g, W_g)$, thus removing the dependency over the number of camera views thereby allowing the model to take an arbitrary number of views as input. 

 \begin{figure}
    \centering
    \includegraphics[width=\linewidth]{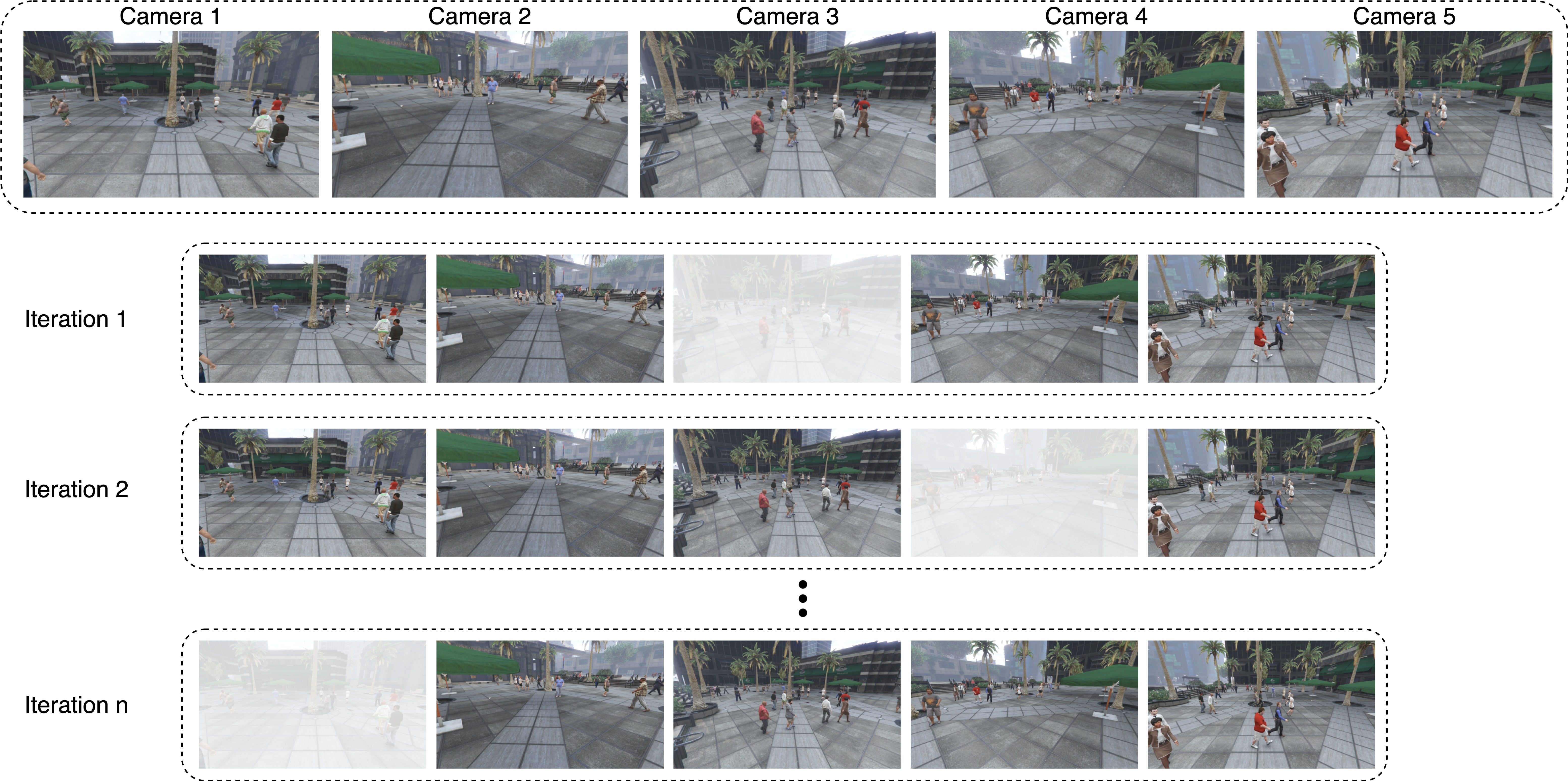}
    \caption{An illustration of our proposed DropView regularization}
    \label{fig:dropview}
\end{figure}

\textbf{DropView Regularization:} Inspired by Dropout~\cite{srivastava2014dropout} as well as work on self-supervised learning which drops color channels to prevent the model from memorization~\cite{jenni2018self,lai2019self}, we propose the DropView regularization technique. For each sample, we randomly select one view to discard during training iterations, as illustrated in Fig~\ref{fig:dropview}. The occupancy map prediction step is done with all the remaining views. We provide a detailed analysis of the effect of this regularization strategy in our experiments.

\textbf{Occupancy Map Prediction:} Similar to MVDet~\cite{hou2020multiview}, we use 3 dilated convolutional layers to predict the occupancy map of size $(H_g,W_g)$. 

\subsection{Loss Function}
The loss function compares the output probabilistic occupancy map $(p)$ with the ground-truth $(g)$. Inspired by the work on saliency estimation in images and vidoes~\cite{bylinskii2018different,Reddy2020TidyingDS,jain2020vinet}, we use the combination of Kullback–Leibler Divergence (KLDiv) and Pearson Cross-Correlation (CC) metrics as a loss function. The final loss function can be written as:
\begin{equation}
L(p,g) = \frac{\sigma(p,g)}{\sigma(p) \times \sigma(g)} - \sum_{i}g_{i} \log \left( \frac{g_{i}}{p_{i}} \right), 
\end{equation} where $\sigma(p,g)$ is the covariance of $p$ and $g$, $\sigma(p)$ is the standard deviation of $p$ and $\sigma(g)$ is the standard deviation of $g$. The loss function was selected empirically using the scene generalization experiment, i.e. training on \multiviewx and testing on \wildtrack, where using KLDiv+CC gave best results (compared with MSE, CC or KLDiv alone).

\begin{table*}[t]
\centering
\caption{Comparison against the \sota methods. Our method refers to the proposed model in Section~\ref{sec:method}. We made five runs for some of the experiments and the variances are presented in the bracket.}
\resizebox{\textwidth}{!}{%
\begin{tabular}{@{}lc|cccc|cccc@{}}
\toprule
\multirow{2}{*}{Method} &
  \multirow{2}{*}{\begin{tabular}[c]{@{}c@{}}ImageNet\\ (pre-train)\end{tabular}} &
  \multicolumn{4}{c|}{\wildtrack} &
  \multicolumn{4}{c}{\multiviewx} \\ \cmidrule(l){3-10} 
                 &     & MODA  & MODP  & Prec  & Recall & MODA  & MODP  & Prec  & Recall \\ \midrule
RCNN  Clustering~\cite{xu2016multi} &  $\times$   & 11.3           & 18.4           & 68.0             & 43.0              & 18.7           & 46.4           & 63.5           & 43.9            \\
POM-CNN~\cite{Fleuret2008MulticameraPT}          &   $\times$  & 23.2           & 30.5           & 75.0             & 55.0              & -              & -              & -              & -               \\
Lopez-Cifuentes \etal~\cite{LpezCifuentes2018SemanticDM}    &  $\times$   & 39.0           & 55.0           & -          & -          & -              & -              & -              & -               \\
Lima \etal~\cite{Lima2021GeneralizableM3}      &  $\times$   & 56.9           & 67.3           & 80.8           & 74.6            & -              & -              & -              & -               \\
DeepMCD~\cite{Chavdarova2017DeepMP}          &  $\times$   & 67.8           & 64.2           & 85.0             & 82.0              & 70.0           & 73.0           & 85.7           & 83.3            \\
Deep-Occlusion \cite{Baqu2017DeepOR}  &  $\times$   & 74.1           & 53.8           & 95.0             & 80.0              & 75.2           & 54.7           & 97.8           & 80.2            \\
MVDet~\cite{hou2020multiview}            &   $\times$  & 88.2           & 75.7           & 94.7           & 93.6            & 83.9           & 79.6           & 96.8           & 86.7            \\
MVDeTr~\cite{hou2021multiview}    &   \checkmark  & 91.5           & 82.1           & 97.4           & 94.0            & 93.7           & 91.3           & 99.5           & 94.2            \\
SHOT~\cite{song2021stacked}    &   $\times$  & 90.2           & 76.5           & 96.1           & 94.0            & 88.3           & 82.0           & 96.6           & 91.5            \\ \bottomrule
Ours             &  $\times$   & 87.2($\pm$0.6) & 74.5($\pm$0.4) & 93.8($\pm$1.6) & 93.4($\pm$1.8)  & 78.6($\pm$0.9) & 78.1($\pm$0.4) & 96.8($\pm$0.5) & 81.3($\pm$0.9)  \\

Ours             & \checkmark & 85.4($\pm$0.4) & {76.7}($\pm$0.2) & {95.2}($\pm$0.4) & 89.9($\pm$0.8)  & 86.9($\pm$0.2) & 79.8($\pm$0.1) & {97.2}($\pm$0.2) & 89.6($\pm$0.2)  \\
Ours (DropView) & \checkmark & 86.7($\pm$0.4) & 76.2($\pm$0.2) & 95.1($\pm$0.3) & 91.4($\pm$0.6) & 88.2($\pm$0.1) & 79.9($\pm$0.0) & 96.8($\pm$0.2) & 91.2($\pm$0.1) \\
\bottomrule
\end{tabular}%
}

\label{tab:sota_table}
\end{table*}

\section{Experiments}
\subsection{Experimental setup}
\textbf{Datasets:} In addition to our proposed GMVD dataset, we use the WildTrack and MultiViewX datasets. The \emph{WildTrack} dataset consists of 7 static calibrated cameras with overlapping fields of view,  covering an area of $12 \times 36 $ $m^2$. The dataset comprises a single 200 second sequence annotated at 2 fps. The image resolution is 1080 $\times$ 1920 pixels. The ground plane grid is discretized into a $480 \times 1440$ grid, where each grid cell is 2.5 $cm$ square. On average, the dataset captures 23.8 persons per frame. The \emph{\multiviewx} dataset is a synthetic dataset which has similar configurations as the \emph{\wildtrack} dataset. However, it consists of 6 static calibrated cameras with overlapping fields of view and 400 synchronized frames of resolution 1080 $\times$ 1920 annotated at 2 fps for ground-truth covering an area of $16 \times 25$ $m^2$. The ground plane grid is discretized into a $640 \times 1000$ grid, where each grid cell is 2.5 $cm$ square. On average, the dataset captures 40 persons per frame. For both datasets, we use the first 90\% frames in training and the last 10\% frames for testing, as done in previous work~\cite{hou2020multiview, Chavdarova2018WILDTRACKAM}.

\textbf{Evaluation metrics:} We use the standard evaluation metrics proposed in \cite{Chavdarova2018WILDTRACKAM}. \emph{Multiple Object Detection Accuracy} (MODA) is the primary performance indicator that accounts for normalized missed detections and false positives, i.e., it considers both false negatives and false positives. \emph{Multiple Object Detection Precision} (MODP) assesses the localization precision \cite{Kasturi2009FrameworkFP}. \emph{Precision} and \emph{Recall} is calculated by Precision = TP/(TP+FP) and Recall = TP/(TP+FN) respectively; where TP, FP and FN are True Positives, False Positives, False Negatives. A threshold of 0.5 meters is used to determine the true positives.

\textbf{State of the Art comparisons:}
We compare against nine different methods. The set includes one monocular object detection baseline (referred to as RCNN clustering~\cite{xu2016multi}); a classical probabilistic occupancy map method~\cite{Fleuret2008MulticameraPT}; four anchor based methods~\cite{Lima2021GeneralizableM3,Baqu2017DeepOR,Chavdarova2017DeepMP,LpezCifuentes2018SemanticDM} and three recent end-to-end trainable deep MVD approaches ~\cite{hou2020multiview,hou2021multiview,song2021stacked}. For generalization experiments, we only compare against the recent \sota methods MVDet~\cite{hou2020multiview}, MVDetr~\cite{hou2021multiview} and SHOT~\cite{song2021stacked}.

\subsection{Implementation Details}

Down sampled images of $720 \times 1,280$ pixels serve as an input to the model. The feature extracted from ResNet-18 has $C=512$ channel features, which is bilinearly interpolated to get the shape of $270 \times 480$. These $(N,C=512,H_f=270,W_f=480)$ extracted features are projected onto top view to obtain $(N,512,H_g,W_g)$ sized features for {\it N} viewpoints, which are average pooled to obtain the ground plane grid shape of $(512,H_g,W_g)$. $H_g$ and $W_g$ vary from scene-to-scene, depending on the area of ground plane. %

The spatial aggregation has three layers of dilated convolution with a $3 \times 3$ kernel size and dilation factor of 1, 2, and 4. Training is done for ten epochs with early stopping; we set batch size as 1, SGD optimizer with momentum = 0.9 has been used with one-cycle learning rate scheduler. A probability of $\tau$ or more on the occupancy grid is considered a detection. For GMVD experiments, $\tau$ is determined using MultiViewX as a validation set, and for other experiments, we use $\tau = 0.4$ in alignment with the previous works. Non-Maximal Suppression (NMS) is applied with a spatial resolution of 0.5m. All training and testing have been performed on a single Nvidia GTX 1080 Ti GPU. Unless specifically mentioned, we always use pre-trained ImageNet~\cite{Deng2009ImageNetAL} weights while training our proposed model.

\begin{table*}[t]
\centering
\caption{Results for evaluating with a varying number of cameras. The model is trained on all 7 cameras on WildTrack, and is tested on 2 different sets of 4 cameras each.}
\resizebox{0.7\linewidth}{!}{%
\begin{tabular}{@{}lcccccccc@{}}
\toprule
      & \multicolumn{4}{c}{Inference on \{1,3,5,7\}} & \multicolumn{4}{c}{Inference on \{2,4,5,6\}} \\ \midrule
        Method  & MODA      & MODP     & Prec     & Recall     & MODA     & MODP     & Prec      & Recall     \\ \midrule
MVDet         & 38.9      & 71.5     & \textbf{93.8}     & 41.6       & 16.2     & 47.6     & 80.3      & 21.4       \\
MVDeTr        & 55.8      & \textbf{76.7}     & 80.8     & 73.2       & 34.6     & 69.2     & 68.6     & 63.8       \\
SHOT          & 66.6      & 75.1     & 91.0     & 73.9       & 46.3     & 67.8     & 88.2      & 53.5       \\
Ours   & 76.5      & 74.0     & 91.7     & 84.0       & \textbf{79.3}     & 71.4     & \textbf{91.1}      & 87.9             \\
Ours (DropView) & \textbf{77.0}      & 74.5     & 90.3     & \textbf{86.2}       & 79.2     & 72.5     & 88.6      & \textbf{90.9}            \\ \bottomrule
\end{tabular}
}

\label{tab:per_inv}
\end{table*}

\subsection{Results}
Like prior works, we evaluate our approach on the WildTrack and MultiViewX datasets in Table~\ref{tab:sota_table}. We find that our proposed models attains satisfactory performance on the test sets of both WildTrack (best MODA score of 87.2)  and MultiViewX (best MODA score of 88.2). This is slightly worse than the recently proposed methods~\cite{hou2021multiview,song2021stacked}, but is far superior to the performance of the classical and the anchor-based MVD methods. However, we would like to highlight that the traditional evaluation protocol is highly misleading since the train and test sets have significant overlap, thereby encouraging overfitting. Therefore, we emphasize the evaluation across a varying number of cameras, changing camera configurations, and on new scenes.

\textbf{Generalization to Varying Number of Cameras:} An interesting scenario that can potentially occur in practical scenarios is the loss of some camera feeds due to various issues. In this case, a model trained with 7 cameras, may need to be able to perform inference with just 4 cameras. To simulate this setting, we train all the models (MVDet, MVDeTr, SHOT and Ours) on all 7 cameras and test them on 2 different sets of 4 cameras (\texttt{\{1,3,5,7\}},\texttt{\{2,4,5,6\}}) in Table~\ref{tab:per_inv}. Our proposed model is able to naturally work in this setting without any issues. For MVDet, MVDeTr, and SHOT, we randomly duplicate 3 of these views to ensure that 7 views are available. We observe that the performance of MVDet, MVDeTr, and SHOT degrades drastically when evaluated in this setting. When trained with the DropView regularization, our proposed model outperforms these methods by a huge margin (MODA of 77.0 vs 66.6 and 79.2 vs 46.3). This experiment clearly illustrates the need for the architectures to automatically work with an arbitrary number of views. Furthermore, since MVDet, MVDeTr, and SHOT learn a separate spatial aggregation module for each view, the spatial aggregation module overfits to the order of input cameras (indicated by the significant performance variations across the two sets). Future works should ensure that the model has permutation invariance to the order of input views in addition to working with an arbitrary number of views.

\begin{table}[t]
\centering
\caption{Scene Generalization : Evaluation of our method while training on synthetic dataset (MultiViewX) and testing on real dataset (WildTrack). Camera 7 of the \wildtrack  dataset was discarded for the experiments in the first five rows.}
\resizebox{\linewidth}{!}{%
\begin{tabular}{@{}lcccccc@{}}
\toprule
Method & \begin{tabular}[c]{@{}c@{}}Inference on\\ total cameras\end{tabular} & \begin{tabular}[c]{@{}c@{}}ImageNet\\ (pre-train)\end{tabular} & MODA  & MODP   & Prec   & Recall \\ \midrule
MVDet & 6 & $\times$   & 17.0   & 65.8 & 60.5 & 48.8 \\
MVDeTr & 6 & \checkmark & 50.2 & 69.1 & 74.0 & 77.3 \\
SHOT   & 6 & $\times$ & 53.6 & 72.0 & 75.2 & 79.8 \\ \midrule
Ours  & 6 & \checkmark & 60.1 & 72.1 & 75.6 & \textbf{88.7} \\ 
Ours (DropView) & 6 & \checkmark & 66.1 & 72.2 & 82.0 & 84.7 \\ 
Ours & 7 & \checkmark & 69.4 & 72.96 & \textbf{83.7} & 86.14 \\
Ours (DropView) & 7 & \checkmark & \textbf{70.7} & \textbf{73.8} & \textbf{89.1} & 80.6 \\\bottomrule
\end{tabular}%
}
\label{tab:scene_general}
\end{table}
\begin{table*}[t]
\centering
\caption{Experiments on the \wildtrack dataset with changing camera configurations}
\resizebox{0.7\linewidth}{!}{%
\begin{tabular}{@{}lllccccccccc@{}}
\toprule
 &  &  & \multicolumn{4}{c}{Inference on \{2,4,5,6\}} & & \multicolumn{4}{c}{Inference on \{1,3,5,7\}} \\ \midrule
 &  & Method  & MODA  & MODP  & Prec & Recall && MODA  & MODP  & Prec & Recall \\ \midrule
\multirow{10}{*}{{\rotatebox[origin=c]{90}{Trained on camera set}}} & 
\multirow{5}{*}{{\rotatebox[origin=c]{90}{\{2,4,5,6\}}}}
& MVDet  & \textbf{85.2} & 72.2 & 92.6 & \textbf{92.5} && 43.2 & 68.2 & \textbf{94.6} & 45.8\\ &
& MVDeTr & 75.4 & \textbf{79.5} & \textbf{96.9} & 77.9 && 41.7 & \textbf{73.7} & 92 & 45.7 \\  & 
& SHOT   & 81.9 & 74.1 & 94.1 & 87.4 && 51.4 & 72.5 & 94.4 & 54.6\\  &              
& Ours   & 81.8 & 73.5 & 93.5 & 87.9 && 66.5 & 71.4 & 94.3 & 70.8\\ &
& Ours (DropView) & 84 & 72.9 & 92.4 & 91.6  && \textbf{75.1} & 71.1 & 94.3 & \textbf{79.9}\\\cmidrule{2-12}
& \multirow{5}{*}{{\rotatebox[origin=c]{90}{\{1,3,5,7\}}}} 
& MVDet     & 27.8 & \textbf{68.7} & \textbf{90.8} & 31.0 && 78.2 & 73.6 & 89.5 & \textbf{88.6} \\  & 
& MVDeTr    & 5.6 & 65.5 & 62.4 & 14.0 && 72.5 & \textbf{78.9} & 95 & 76.5  \\  &            
& SHOT      & 15.3 & 62.9 & 89.2 & 17.4 && 79.7 & 76.4 & \textbf{95.7} & 83.5   \\ & 
& Ours     & 52.4 & 67.4 & 81 & 68.5 && 76.4 & 74.6 & 91.5 & 84.1  \\ &
& Ours (DropView) & \textbf{62.6} & 67.4 & 86.7 & \textbf{73.9}  && \textbf{80.8} & 74.0 & 94.2 & 86 \\ \midrule
\end{tabular}
}

\label{tab:camera_config}
\end{table*}

\begin{table}[t]
\centering
\caption{Changing configuration and scene generalization experiment on the setting introduced in~\cite{song2021stacked}}
\footnotesize
\begin{tabular}{@{}lcccc@{}}
\toprule
Method   & MODA & MODP & Prec & Recall \\ \midrule
MVDet           & 33.0 & 76.5 & 64.5      & 73.4   \\
MVDeTr          & 56.5 & 70.8 & 85.0      & 68.6       \\ 
SHOT            & 49.1 & \textbf{77.0} & 73.3      & \textbf{77.1}   \\
Ours            & 57.8 & 76.5 & 88.7 & 66.3   \\
Ours (DropView) & \textbf{66.1} & 75.8  & \textbf{89.3}  &  75.2       \\ \bottomrule
\end{tabular}

\label{tab:shot_exp}
\end{table}

\begin{table}[t]
\centering
\caption{Evaluation when trained on GMVD training set: first row shows the result on GMVD test set and second row is when tested on \wildtrack dataset. }
\small
\begin{tabular}{@{}lcccc@{}}
\toprule
Inference on & MODA & MODP & Prec & Recall \\ \midrule
GMVD & 68.2 & 76.3 & 91.5  & 75.5   \\
WildTrack & 80.1 & 75.6 & 90.9  & 89.1      \\ \bottomrule
\end{tabular}

\label{tab:gmvd_dataset_exp}
\end{table}
\textbf{Generalization to New Camera Configurations:} Another practical scenario that we explore is when the camera positions are varied between the train and test sets. We train all the models on two sets of camera views and then test the trained models on both sets. The results are provided in Table~\ref{tab:camera_config}. When the models are evaluated on the same camera configuration, all the models have satisfactory performance. However, when evaluated on the different camera configuration, MVDet, MVDeTr, and SHOT see a huge degradation in performance. Our model is fairly robust to the changing camera configuration. Especially when trained with DropView regularization, the resulting model outperforms all other models by over 20 percentage points.

\textbf{Scene Generalization:} Finally, an important concern with the practical utility of MVD methods is that since real-world data is scarce, a trained model should be able to generalize to new scenes. We first evaluate the scene generalization abilities of the MVD methods by training them on MultiViewX and evaluating them on WildTrack in Table~\ref{tab:scene_general}. Our proposed model is able to utilize the extra camera present in the WildTrack dataset and achieves a MODA score of 70.7. This further highlights the benefits of an architecture that works with arbitrary number of views, since the performance during inference can be enhanced by adding more view. However, even without the additional view, our model achieves a MODA score of 66.1, which is much higher than SHOT which only achieves a MODA score of 53.6. 

In addition to this, we perform the scene generalization experiment proposed in \cite{song2021stacked} where the MultiViewX scene is split into two halves, and each half is covered using 3 cameras each. In this setting as well (Table~\ref{tab:shot_exp}), our proposed approach with DropView regularization has a MODA score of 66.1, which is significantly higher than both SHOT (49.1) and MVDeTr (56.5).

\textbf{GMVD Benchmark:} Having shown that our proposed model is capable of comprehensive generalization abilities, we benchmark our proposed approach on the GMVD dataset (Table~\ref{tab:gmvd_dataset_exp}). We train our model on the training set of the GMVD dataset and use MultiViewX dataset for validation. Since each sequence in the training set has a different number of cameras, \emph{none} of the existing methods can be adapted to this setting, since they can be trained only on a \emph{fixed} set of cameras. 
When evaluated on WildTrack, our model is able to achieve a MODA score of 80.1, which is a significant improvement over the results from training on MultiViewX. Notably, this shows that training on our synthetic dataset, we can nearly attain the same performance as training on WildTrack itself. When evaluated on GMVD test set, our model achieves a MODA score of 68.2. The results empirically suggest the difficulty of the GMVD test set, compared to WildTrack and MultiViewX, resulting from a distinct train-test split and the presence of extensive variations. We believe that our dataset can serve two important purposes. The first is as a diverse, synthetic dataset from which a model can be adapted to real-world data. The second is that the GMVD dataset itself can be a challenging benchmark to evaluate the generalization capabilities of MVD methods. In this setting, MultiViewX being used for validation is ideal, since this ensures that no information from the test set is leaked during training. 

\section{Discussion and Future work}
The biggest limitation in the field of Multi-View Detection is that real-world capture of data is extremely challenging due to the difficulty in collecting a dataset with people in addition to the challenges involved in the hardware setup and annotations. The absence of a large, diverse benchmark significantly hampers the progress of this topic. Therefore, the existing WildTrack dataset is extremely valuable for the community. However, due to its limited size and variety, it is not suitable for training and should only be used to evaluate the generalization abilities of the models. In this regard, we hope that our proposed dataset and our barebone model serves as a useful tool in bridging the gap between the theory and real-world application of MVD methods. In our work, we have not explored the use of unsupervised domain adaptation techniques to bridge the gap between the feature distributions of the synthetic and real datasets and the direction is left for exploration in the future work.

\section{Conclusion}
We find the current Multi-View Detection setup severely limited and encouraging models to overfit the training configuration. Therefore, we conceptualize and propose novel experimental setups to evaluate the generalization capabilities of MVD models in a more practical setting. We find the state-of-the-art models to have poor generalization capabilities on our proposed setups. To alleviate this issue, we introduce changes to the feature aggregation strategy, loss function, as well as a novel regularization strategy. With the help of comprehensive experiments, we demonstrate the benefits of our proposed architecture. In addition to this, we propose a diverse, synthetic, but realistic dataset which can be used both as an evaluation benchmark, as well as a training dataset for various MVD methods. Overall, we hope our work plays a crucial role in steering the community towards more practical Multi-View Detection solutions. 

\bibliographystyle{IEEEtran}
\bibliography{IEEEabrv,egbib}

\begin{thebibliography}{10}
\providecommand{\url}[1]{#1}
\csname url@rmstyle\endcsname
\providecommand{\newblock}{\relax}
\providecommand{\bibinfo}[2]{#2}
\providecommand\BIBentrySTDinterwordspacing{\spaceskip=0pt\relax}
\providecommand\BIBentryALTinterwordstretchfactor{4}
\providecommand\BIBentryALTinterwordspacing{\spaceskip=\fontdimen2\font plus
\BIBentryALTinterwordstretchfactor\fontdimen3\font minus
  \fontdimen4\font\relax}
\providecommand\BIBforeignlanguage[2]{{%
\expandafter\ifx\csname l@#1\endcsname\relax
\typeout{** WARNING: IEEEtran.bst: No hyphenation pattern has been}%
\typeout{** loaded for the language `#1'. Using the pattern for}%
\typeout{** the default language instead.}%
\else
\language=\csname l@#1\endcsname
\fi
#2}}

\bibitem{Fleuret2008MulticameraPT}
F.~Fleuret, J.~Berclaz, R.~Lengagne, and P.~Fua, ``Multicamera people tracking
  with a probabilistic occupancy map,'' \emph{IEEE Transactions on Pattern
  Analysis and Machine Intelligence}, vol.~30, pp. 267--282, 2008.

\bibitem{berclaz2011multiple}
J.~Berclaz, F.~Fleuret, E.~Turetken, and P.~Fua, ``Multiple object tracking
  using k-shortest paths optimization,'' \emph{IEEE transactions on pattern
  analysis and machine intelligence}, vol.~33, no.~9, pp. 1806--1819, 2011.

\bibitem{alahi2011sparsity}
A.~Alahi, L.~Jacques, Y.~Boursier, and P.~Vandergheynst, ``Sparsity driven
  people localization with a heterogeneous network of cameras,'' \emph{Journal
  of Mathematical Imaging and Vision}, vol.~41, no.~1, pp. 39--58, 2011.

\bibitem{kong2020foveabox}
T.~Kong, F.~Sun, H.~Liu, Y.~Jiang, L.~Li, and J.~Shi, ``Foveabox: Beyound
  anchor-based object detection,'' \emph{IEEE Transactions on Image
  Processing}, vol.~29, pp. 7389--7398, 2020.

\bibitem{hou2020multiview}
Y.~Hou, L.~Zheng, and S.~Gould, ``Multiview detection with feature perspective
  transformation,'' in \emph{ECCV}, 2020.

\bibitem{hou2021multiview}
Y.~Hou and L.~Zheng, ``Multiview detection with shadow transformer (and
  view-coherent data augmentation),'' in \emph{Proceedings of the 29th ACM
  International Conference on Multimedia}, 2021, pp. 1673--1682.

\bibitem{song2021stacked}
L.~Song, J.~Wu, M.~Yang, Q.~Zhang, Y.~Li, and J.~Yuan, ``Stacked homography
  transformations for multi-view pedestrian detection,'' in \emph{Proceedings
  of the IEEE/CVF International Conference on Computer Vision}, 2021, pp.
  6049--6057.

\bibitem{Chavdarova2018WILDTRACKAM}
T.~Chavdarova, P.~Baqu{\'e}, S.~Bouquet, A.~Maksai, C.~Jose, T.~M. Bagautdinov,
  L.~Lettry, P.~Fua, L.~Gool, and F.~Fleuret, ``Wildtrack: A multi-camera hd
  dataset for dense unscripted pedestrian detection,'' \emph{2018 IEEE/CVF
  Conference on Computer Vision and Pattern Recognition}, pp. 5030--5039, 2018.

\bibitem{Roig2011ConditionalRF}
G.~Roig, X.~Boix, H.~B. Shitrit, and P.~Fua, ``Conditional random fields for
  multi-camera object detection,'' \emph{2011 International Conference on
  Computer Vision}, pp. 563--570, 2011.

\bibitem{Ren2015FasterRT}
S.~Ren, K.~He, R.~B. Girshick, and J.~Sun, ``Faster r-cnn: Towards real-time
  object detection with region proposal networks,'' \emph{IEEE Transactions on
  Pattern Analysis and Machine Intelligence}, vol.~39, pp. 1137--1149, 2015.

\bibitem{Liu2016SSDSS}
W.~Liu, D.~Anguelov, D.~Erhan, C.~Szegedy, S.~E. Reed, C.-Y. Fu, and A.~Berg,
  ``Ssd: Single shot multibox detector,'' in \emph{ECCV}, 2016.

\bibitem{Redmon2016YouOL}
J.~Redmon, S.~Divvala, R.~B. Girshick, and A.~Farhadi, ``You only look once:
  Unified, real-time object detection,'' \emph{2016 IEEE Conference on Computer
  Vision and Pattern Recognition (CVPR)}, pp. 779--788, 2016.

\bibitem{xu2016multi}
Y.~Xu, X.~Liu, Y.~Liu, and S.-C. Zhu, ``Multi-view people tracking via
  hierarchical trajectory composition,'' in \emph{Proceedings of the IEEE
  Conference on Computer Vision and Pattern Recognition}, 2016, pp. 4256--4265.

\bibitem{Baqu2017DeepOR}
P.~Baqu{\'e}, F.~Fleuret, and P.~Fua, ``Deep occlusion reasoning for
  multi-camera multi-target detection,'' \emph{2017 IEEE International
  Conference on Computer Vision (ICCV)}, pp. 271--279, 2017.

\bibitem{Chavdarova2017DeepMP}
T.~Chavdarova and F.~Fleuret, ``Deep multi-camera people detection,''
  \emph{2017 16th IEEE International Conference on Machine Learning and
  Applications (ICMLA)}, pp. 848--853, 2017.

\bibitem{He2016DeepRL}
K.~He, X.~Zhang, S.~Ren, and J.~Sun, ``Deep residual learning for image
  recognition,'' \emph{2016 IEEE Conference on Computer Vision and Pattern
  Recognition (CVPR)}, pp. 770--778, 2016.

\bibitem{zhu2020deformable}
X.~Zhu, W.~Su, L.~Lu, B.~Li, X.~Wang, and J.~Dai, ``Deformable detr: Deformable
  transformers for end-to-end object detection,'' \emph{arXiv preprint
  arXiv:2010.04159}, 2020.

\bibitem{gong2021mdalu}
R.~Gong, D.~Dai, Y.~Chen, W.~Li, and L.~Van~Gool, ``mdalu: Multi-source domain
  adaptation and label unification with partial datasets,'' in
  \emph{Proceedings of the IEEE/CVF International Conference on Computer
  Vision}, 2021, pp. 8876--8885.

\bibitem{Zhao_2021_CVPR}
Y.~Zhao, Z.~Zhong, F.~Yang, Z.~Luo, Y.~Lin, S.~Li, and N.~Sebe, ``Learning to
  generalize unseen domains via memory-based multi-source meta-learning for
  person re-identification,'' in \emph{Proceedings of the IEEE/CVF Conference
  on Computer Vision and Pattern Recognition (CVPR)}, June 2021, pp.
  6277--6286.

\bibitem{scripthook}
A.~Blade, ``{Script Hook V},'' \url{http://www.dev-c.com/gtav/scripthookv/},
  2008, [Online; accessed 19-July-2008].

\bibitem{Kohl_2020_CVPR_Workshops}
P.~Kohl, A.~Specker, A.~Schumann, and J.~Beyerer, ``The mta dataset for
  multi-target multi-camera pedestrian tracking by weighted distance
  aggregation,'' in \emph{The IEEE/CVF Conference on Computer Vision and
  Pattern Recognition (CVPR) Workshops}, June 2020.

\bibitem{sun2019dissecting}
X.~Sun and L.~Zheng, ``Dissecting person re-identification from the viewpoint
  of viewpoint,'' in \emph{CVPR}, 2019.

\bibitem{srivastava2014dropout}
N.~Srivastava, G.~Hinton, A.~Krizhevsky, I.~Sutskever, and R.~Salakhutdinov,
  ``Dropout: a simple way to prevent neural networks from overfitting,''
  \emph{The journal of machine learning research}, vol.~15, no.~1, pp.
  1929--1958, 2014.

\bibitem{jenni2018self}
S.~Jenni and P.~Favaro, ``Self-supervised feature learning by learning to spot
  artifacts,'' in \emph{Proceedings of the IEEE Conference on Computer Vision
  and Pattern Recognition}, 2018, pp. 2733--2742.

\bibitem{lai2019self}
Z.~Lai and W.~Xie, ``Self-supervised learning for video correspondence flow,''
  \emph{arXiv preprint arXiv:1905.00875}, 2019.

\bibitem{bylinskii2018different}
Z.~Bylinskii, T.~Judd, A.~Oliva, A.~Torralba, and F.~Durand, ``What do
  different evaluation metrics tell us about saliency models?'' \emph{IEEE
  transactions on pattern analysis and machine intelligence}, vol.~41, no.~3,
  pp. 740--757, 2018.

\bibitem{Reddy2020TidyingDS}
N.~Reddy, S.~Jain, P.~Yarlagadda, and V.~Gandhi, ``Tidying deep saliency
  prediction architectures,'' \emph{2020 IEEE/RSJ International Conference on
  Intelligent Robots and Systems (IROS)}, pp. 10\,241--10\,247, 2020.

\bibitem{jain2020vinet}
S.~Jain, P.~Yarlagadda, S.~Jyoti, S.~Karthik, R.~Subramanian, and V.~Gandhi,
  ``Vinet: Pushing the limits of visual modality for audio-visual saliency
  prediction,'' \emph{arXiv preprint arXiv:2012.06170}, 2020.

\bibitem{LpezCifuentes2018SemanticDM}
A.~L{\'o}pez-Cifuentes, M.~Escudero-Vi{\~n}olo, J.~Besc{\'o}s, and
  P.~Carballeira, ``Semantic driven multi-camera pedestrian detection,''
  \emph{ArXiv}, vol. abs/1812.10779, 2018.

\bibitem{Lima2021GeneralizableM3}
J.~Lima, R.~Roberto, L.~Figueiredo, F.~Sim{\~o}es, and V.~Teichrieb,
  ``Generalizable multi-camera 3d pedestrian detection,'' \emph{ArXiv}, vol.
  abs/2104.05813, 2021.

\bibitem{Kasturi2009FrameworkFP}
R.~Kasturi, D.~Goldgof, P.~Soundararajan, V.~Manohar, J.~S. Garofolo,
  R.~Bowers, M.~Boonstra, V.~Korzhova, and J.~Zhang, ``Framework for
  performance evaluation of face, text, and vehicle detection and tracking in
  video: Data, metrics, and protocol,'' \emph{IEEE Transactions on Pattern
  Analysis and Machine Intelligence}, vol.~31, pp. 319--336, 2009.

\bibitem{Deng2009ImageNetAL}
J.~Deng, W.~Dong, R.~Socher, L.-J. Li, K.~Li, and L.~Fei-Fei, ``Imagenet: A
  large-scale hierarchical image database,'' in \emph{CVPR}, 2009.

\end{thebibliography}

\clearpage
\appendix

\begin{figure}[t]
    \centering
    \includegraphics[scale=0.25]{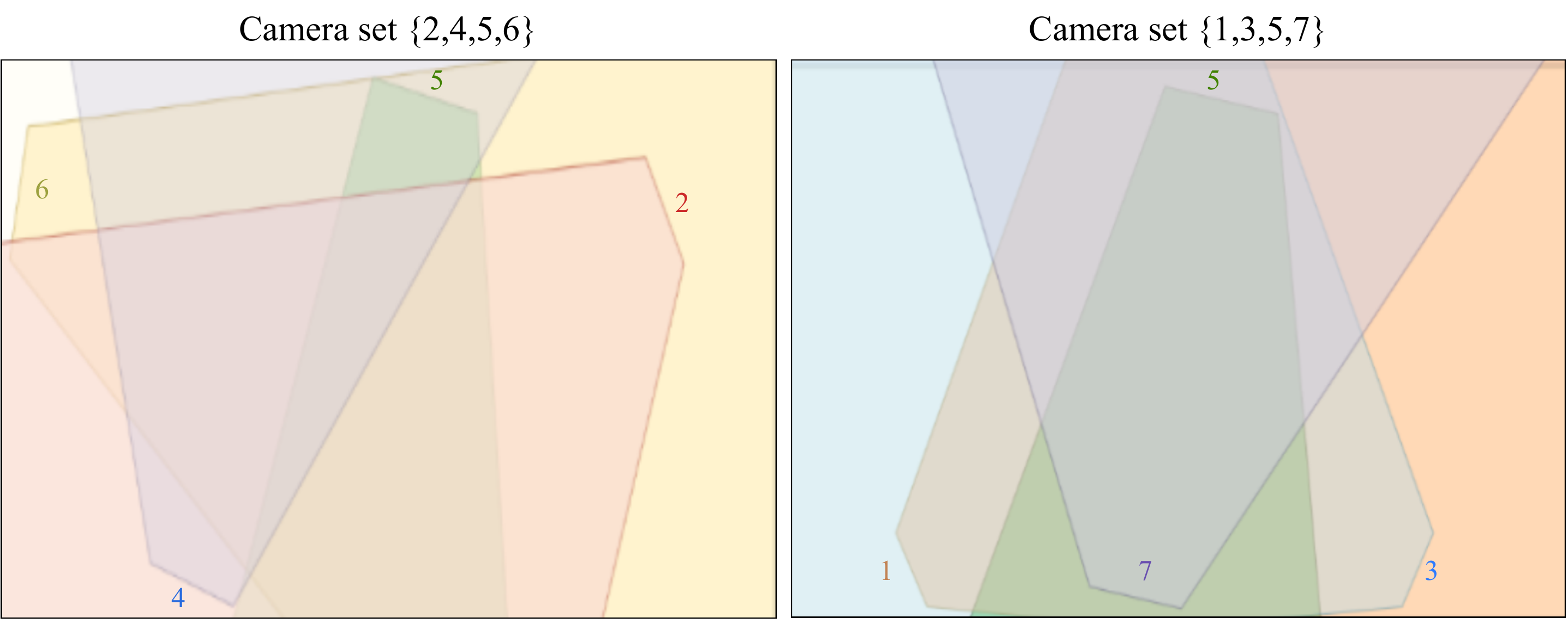}
    \caption{Camera splits of \wildtrack dataset for changing camera configuration experiment.}
    \label{fig:wildtrack_cam_split}
\end{figure}

\begin{figure*}[t]
    \centering
    \includegraphics[width=\linewidth]{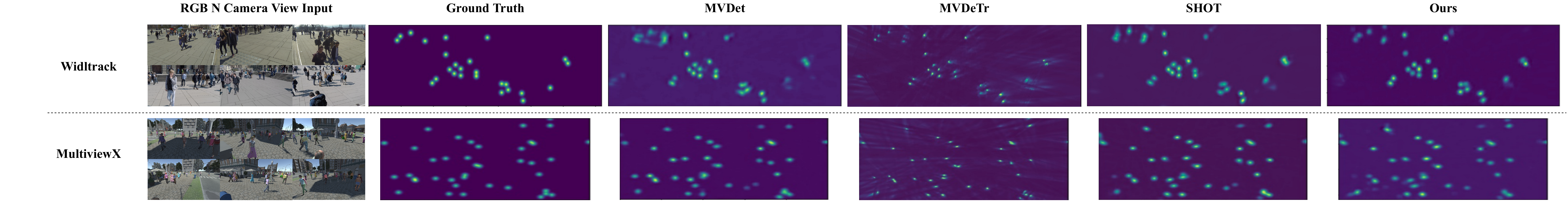}
    \caption{ Sample frames from \wildtrack and \multiviewx dataset with corresponding occupancy maps of ground truth, our result MVDet, MVDeTr and SHOT for comparison. We can see the clusters forming in the MVDet predictions, in contrast our method gives much sharper and distinct predictions.}
    \label{fig:wildtrack_results}
\end{figure*}

\begin{figure*}[t]
    \centering
    \includegraphics[width=\linewidth]{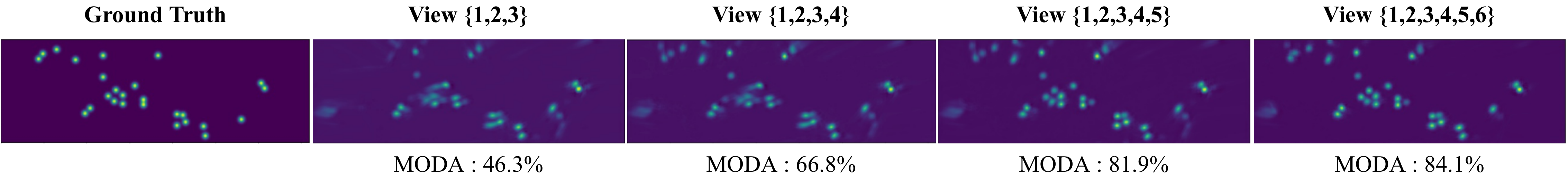}
    \caption{Occupancy maps for varying number of cameras on \wildtrack dataset when trained on seven cameras and tested on varying subsets of the cameras.}
    \label{fig:wildtrack_varying_cam}
\end{figure*}

\section{Choice of Loss Function}
\label{sec:choice_loss}

\begin{table}[htbp!]
\begin{center}
\resizebox{\linewidth}{!}{%
\begin{tabular}{@{}cccccc@{}}
\toprule
\textbf{Method} & \textbf{\begin{tabular}[c]{@{}c@{}}ImageNet\\ (pre-train)\end{tabular}} & \textbf{MODA}  & \textbf{MODP}   & \textbf{Prec}   & \textbf{Recall} \\ \midrule
 MSE & \checkmark & 57.3($\pm$0.2) & 72.6($\pm$0.0) & 75.6($\pm$0.1) & 84.5($\pm$0.05) \\
 CC & \checkmark & 55.5($\pm$5.5) & \textbf{74.2}($\pm$0.4) & 72.1($\pm$4.4) & \textbf{89.5}($\pm$2.6) \\
 KL & \checkmark & 62.5($\pm$0.1) & 73.4($\pm$0.04) & \textbf{89.1}($\pm$0.0) & 71.3($\pm$0.0) \\
 KLCC & \checkmark & \textbf{69.4}($\pm$0.6) & \ 72.96($\pm$0.2) & 83.74($\pm$0.5) & 86.14($\pm$0.3) \\ \bottomrule
\end{tabular}%
}
\end{center}
\caption{Choice of Loss Function: we present an ablation study for our proposed method on the scene generalization experiment. Overall, the model trained with both KL-Divergence and Cross-Correlation achieves the best performance.}
\label{tab:scene_loss_compare}
\end{table}

We ablate the choice of the loss function in Table~\ref{tab:scene_loss_compare} for the scene generalization experiment. We consider the Mean Squared Error (MSE), KL-Divergence(KL), Pearson Cross-Correlation (CC), as well as our chosen loss function (KL+CC). We find that the combination of KL-Divergence and Pearson Cross-Correlation achieves significantly better results than any other loss function. 

\section{Qualitative results}
\label{sec:qualitative_results}

First we show the predicted occupancy maps of MVDet, MVDeTr, SHOT and our method and compare them with the ground truth, in the traditional setting. Subsequently, qualitative results are shown w.r.t to three generalization abilities obtained from both the \wildtrack and \multiviewx datasets.
\subsection{\wildtrack Dataset}
The traditionally evaluated results which contains occupancy maps of ground truth, our method, MVDet, MVDeTr and SHOT are shown in Fig. \ref{fig:wildtrack_results}. The occupancy map from our method which uses average pooling, KLCC loss function and ImageNet pretraining gives us more accurate localization as compared to the base MVDet architecture. The results (maps) are competitive when compared to SHOT and MVDeTr. The maps obtained using MVDeTr are sharper and focused, however, it also has more false positives.

\textbf{Varying number of cameras:} The output occupancy map for varying number of cameras are shown in Fig. \ref{fig:wildtrack_varying_cam}. \wildtrack consists of seven cameras, we show the results inferred with three cameras upto six cameras. As the number of views are increasing, we get an accurately localized occupancy map.

\begin{figure*}[t]
    \centering
    \includegraphics[width=0.98\linewidth]{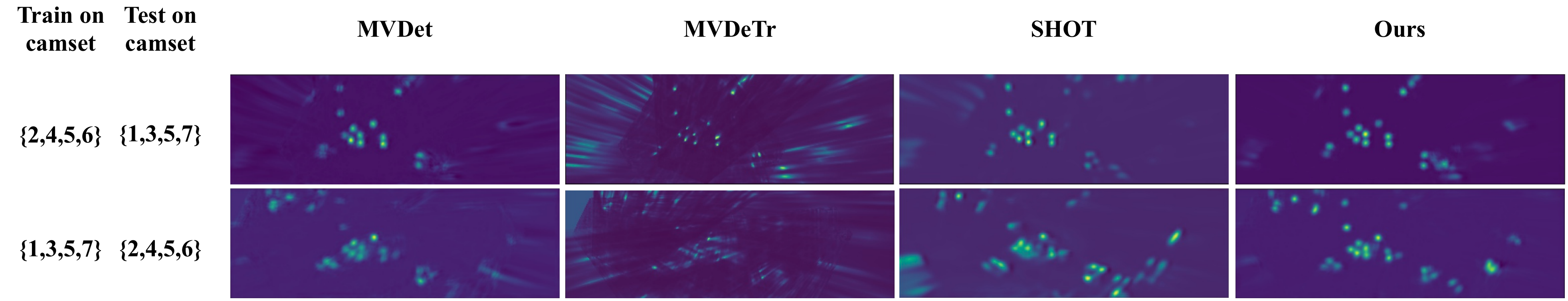}
    \caption{Result occupancy maps for cross subset evaluation from \wildtrack dataset.}
    \label{fig:wildtrack_cam_subset}
\end{figure*}

\textbf{Changing camera configurations:} The output occupancy map for cross subset evaluation are shown in Fig. \ref{fig:wildtrack_cam_subset}. Here, we have the occupancy maps for a model trained on one set and tested on other set. For example, trained on camera views one, three, five and seven and tested on cameras two, four, five and six or vice-versa like the camera splits shown in Figure \ref{fig:wildtrack_cam_split}. Clearly the pre-training is improving localization in both the methods. Furthermore, our method with average pooling is better at disambiguating the occlusions and also giving brighter outputs (resulting in sharp maxima's).

\subsection{\multiviewx Dataset}
In this subsection the qualitative results for \multiviewx dataset are been shown. We consider similar configurations as in the Wildtrack dataset. The obtained results clearly indicates the improvements our method brings over the MVDet, MVDeTr and SHOT model and observations are similar to that of the Wildtrack dataset.  Fig. \ref{fig:wildtrack_results} shows the traditionally evaluated results.

\begin{figure*}[t]
    \centering
    \includegraphics[width=0.7\linewidth]{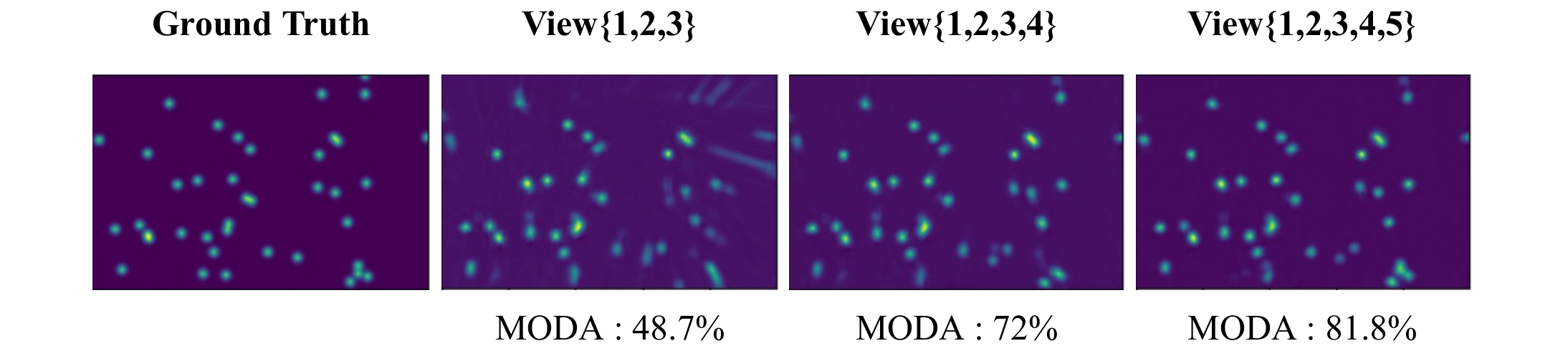}
    \caption{Occupancy maps for varying number of cameras on \multiviewx dataset when trained on seven cameras and tested on varying subsets of the cameras.}
    \label{fig:mx_varying_cam}
\end{figure*}

\textbf{Varying number of cameras:} The output occupancy map for varying number of cameras are shown in Fig. \ref{fig:mx_varying_cam}. \multiviewx consists of six cameras, we show the results inferred with three cameras upto five cameras. As the number of views are increasing, we get an accurately localized occupancy map. 

\begin{figure*}[t!]
    \centering
    \includegraphics[scale=0.2]{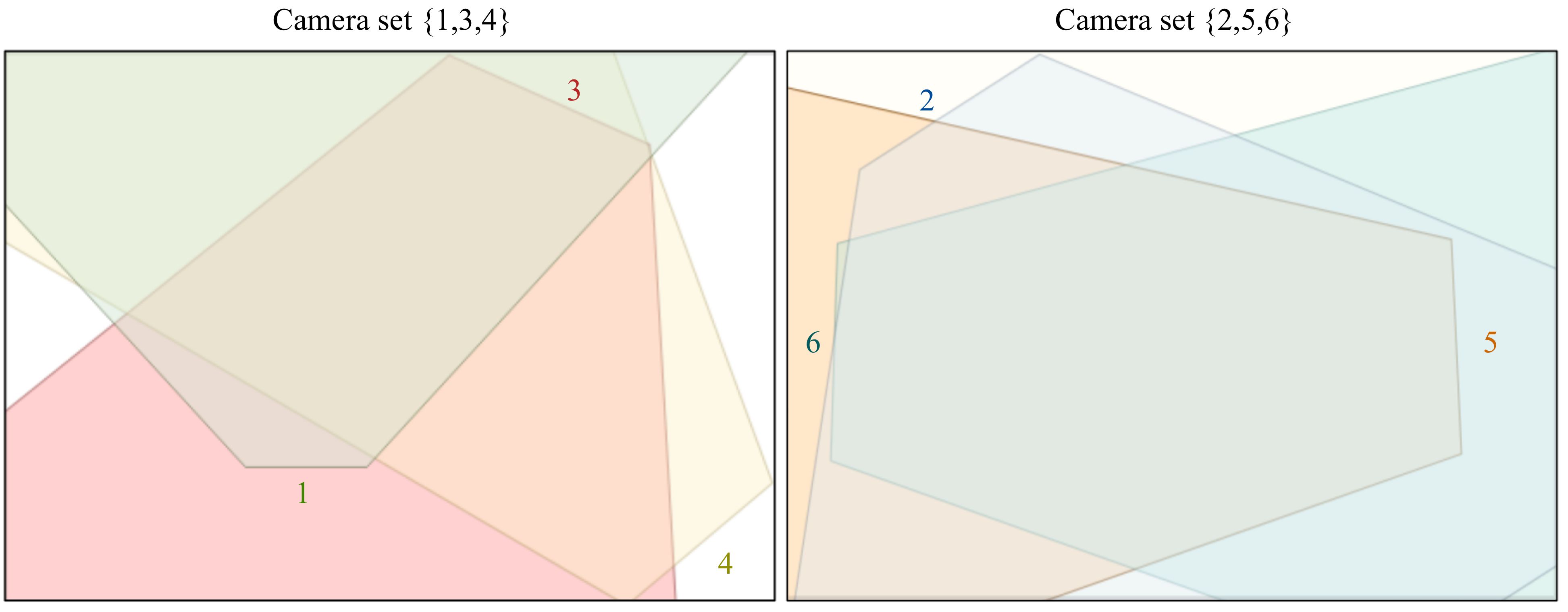}
    \caption{Camera splits of \multiviewx dataset for changing camera configuration experiment shown in Table \ref{tab:camera_config_supp}.}
    \label{fig:mx_cam_split}
\end{figure*}

\begin{figure*}[t!]
    \centering
    \includegraphics[scale=0.5]{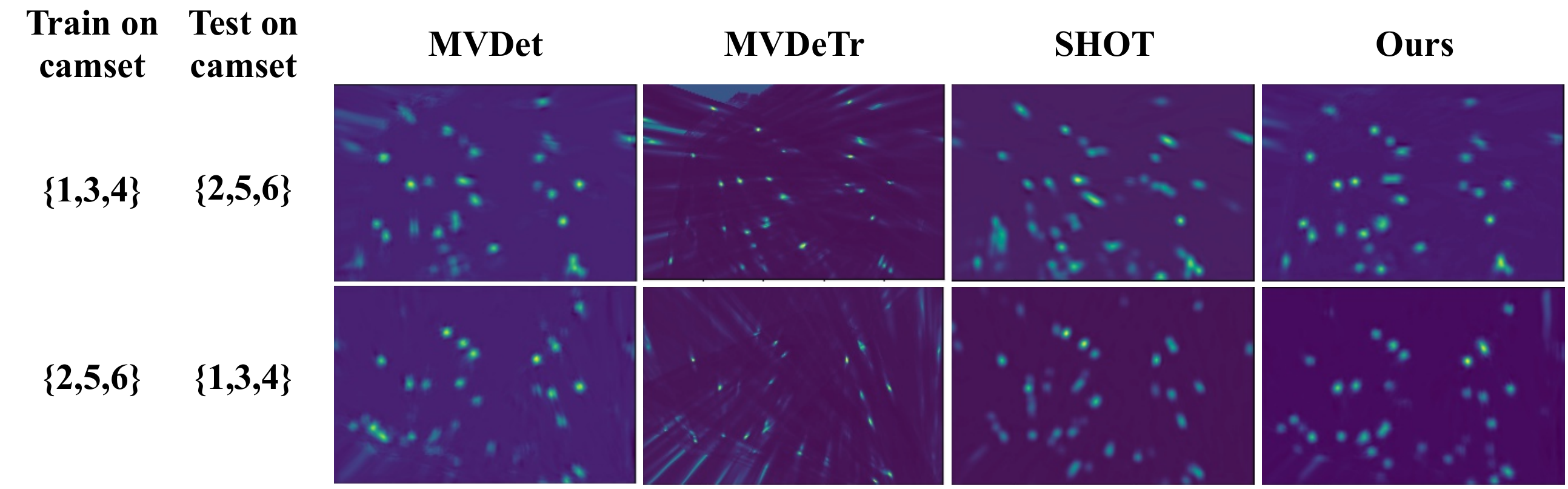}
    \caption{Result occupancy maps for cross subset evaluation from \wildtrack dataset.}
    \label{fig:mx_cam_subset}
\end{figure*}

\textbf{Changing camera configurations:} The output occupancy map for cross subset evaluation are shown in Fig. \ref{fig:mx_cam_subset}. Here, we have the occupancy maps for a model trained on one set and tested on other set. For example, trained on camera views one, three, and four and tested on cameras two, five and six or vice-versa, the camera splits are shown in Figure \ref{fig:mx_cam_split} and their results are shown in Table \ref{tab:camera_config_supp}.

\begin{table}[h!]
\centering
\resizebox{\linewidth}{!}{%
\begin{tabular}{@{}lllccccccccc@{}}
\toprule
 &  &  & \multicolumn{4}{c}{Inference on \{1,3,4\}} & & \multicolumn{4}{c}{Inference on \{2,5,6\}} \\ \midrule
 &  & Method  & MODA  & MODP  & Prec & Recall && MODA  & MODP  & Prec & Recall \\ \midrule
\multirow{10}{*}{{\rotatebox[origin=c]{90}{Trained on camera set}}} & 
\multirow{5}{*}{{\rotatebox[origin=c]{90}{\{1,3,4\}}}}
& MVDet  & 72 & 76.1 & 93.5 & 77.4 && 46.3 &66.4 &94.5 &49.1\\ &
& MVDeTr & \textbf{77.4} & \textbf{85.1} & 97.9 & 79 && 60.4 & 71.3 & 95.4 & 63.5\\  & 
& SHOT   & 74.3 & 76.3 & 94.1 & \textbf{79.3} && 37.3 & 67 & 67.5 & \textbf{72.1}\\  &
& Ours   & 67.7 & 76.4 & 96.2 & 70.5 && 59.6 & 73.4 & 94.7 & 63.2 \\ &
& Ours (DropView) & 67.3 & 75.3 & \textbf{98.4} & 68.5 && \textbf{62.9} & \textbf{73.6} & \textbf{96.3} & 65.4\\\cmidrule{2-12}
& \multirow{5}{*}{{\rotatebox[origin=c]{90}{\{2,5,6\}}}} 
& MVDet     & 34.3 & 66.2 & 93.8 & 36.7 && 77.6 & 77.4 & 93.8 & 83.1 \\  & 
& MVDeTr    & 51.1 & 72.1 & \textbf{94.9} & 54 && \textbf{83.1} & \textbf{87.1} & \textbf{97.8} & \textbf{85}\\  &            
& SHOT      & 47.3 & \textbf{73} & 94.2 & 50.3 && 80.7 & 78.7 & 96.1 & 84.1\\ & 
& Ours     & 45.8 & 71.8 & 94.5 & 48.6 && 76.1 & 78.7 & 95.9 & 79.5 \\ &
& Ours (DropView) & \textbf{53.4} & 71.6 & 88.2 & \textbf{61.6} && 75.2 & 77.4 & 92.8 & 81.5\\ \midrule
\end{tabular}
}
\caption{Experiments on the \multiviewx dataset with changing camera configurations}
\label{tab:camera_config_supp}
\end{table}

\subsection{Scene Generalization}
\begin{figure*}[t!]
    \centering
    \includegraphics[width=\linewidth]{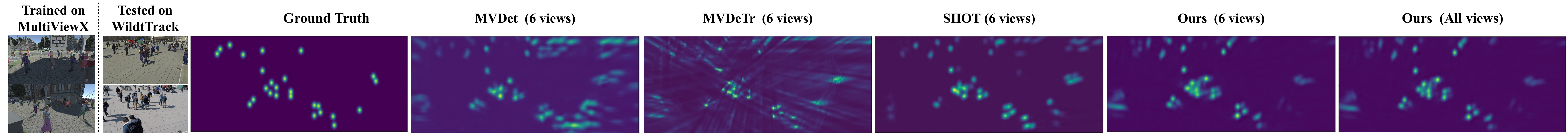}
    \caption{Occupancy maps obtained on inference from \wildtrack dataset where the models where trained on the synthetic dataset (\multiviewx).}
    \label{fig:scene_general}
\end{figure*}

The qualitative results of output occupancy map for cross-dataset evaluation are shown in Fig. \ref{fig:scene_general}, when we train on synthetic dataset (\multiviewx) and test on real dataset (\wildtrack).
First four occupancy maps are the outputs of MVDet, MVDeTr, SHOT and our method when tested on only 6 views of \wildtrack dataset for having a fair comparison with other methods. We also show the output occupancy map when tested on all the views of \wildtrack dataset. Our method provides accurately localized occupancy maps and disambiguate the occlusions as compared to other methods.

\end{document}